\documentclass{article}

\usepackage[preprint]{corl_2026} 
\usepackage{amsmath}
\usepackage{amssymb}
\usepackage{graphicx}
\usepackage{booktabs}

\title{Where Should RL Post-Training Compute Go?\\
Model Size, Search, Learning, and Feedback}

\author{
  Patrick Wilhelm\\
  BIFOLD\\
  Technische Universität Berlin \\
  Germany\\
  \texttt{patrick.wilhelm@tu-berlin.de} \\
  \And
  Odej Kao \\
  BIFOLD\\
  Technische Universität Berlin \\
  Germany\\
}

\begin{document}
\maketitle


\begin{abstract}
Reinforcement Learning (RL) post-training is increasingly used to adapt foundation models for reasoning, planning, and feedback-driven robot-learning pipelines, but constrained post-training resources are often summarized by a single total FLOP budget.
We study the fixed-budget decision problem behind this practice: under the same post-training budget, should one use a larger policy, train a smaller policy longer, generate more rollout search, or spend compute on stronger reward feedback?
We introduce a FLOP-accounting framework for GRPO post-training that decomposes compute into rollout/search, policy-update/learning, and reward- or feedback-model evaluation.
Across LoRA-adapted Qwen2.5 policies, we find conditional allocation frontiers: the best observed allocation changes with model size, compute budget, reward system, and evaluation target.
Same-FLOP model-size comparisons show that model choice and training allocation are coupled because larger policies consume more per-token compute and therefore buy fewer updates or rollouts under the same budget.
Reward systems also change the accounting: rule-based rewards spend nearly all non-update compute on policy rollouts, while PRM-style feedback allocates a visible part of the budget to reward-model inference.
We present RACE as a diagnostic pilot-grid protocol, not a guarantee of held-out improvement, for identifying allocation regimes before expensive validation runs; our results suggest that RL post-training papers should report total FLOPs together with how compute is divided among model size, search, learning, and feedback.
\end{abstract}



\section{Introduction}

Robot-learning systems increasingly use foundation models for language-conditioned planning, feedback generation, reward modeling, high-level reasoning, and vision-language-action policies \citep{ahn2022saycan,huang2023inner,driess2023palme,brohan2022rt1,brohan2023rt2,openx2023,kim2024openvla}. Improving these components often requires reinforcement learning (RL) post-training, but current scaling analyses usually summarize its cost by total FLOPs. We study the compute-allocation problem behind RL post-training of reasoning and feedback modules for robot-learning systems. That problem asks a central systems question: under a fixed post-training budget, should compute be spent on a larger policy, a smaller policy trained longer, more rollout search, more policy updates, or more expensive reward feedback?

This question is different from pretraining scaling. Pretraining primarily allocates compute between parameters and data \citep{kaplan2020scaling, hoffmann2022training}; RL post-training also generates trajectories, scores them, and updates a policy. Two runs with the same total compute can therefore behave differently if one spends compute on diverse rollouts while another spends it on repeated updates from fewer trajectories. At the same time, larger policies consume more per-token compute, so the same FLOP budget buys fewer rollouts and fewer update steps. The issue becomes more pronounced when feedback comes from learned verifiers or process reward models, whose inference cost competes with policy rollout and update compute.

This makes total-compute reporting under-specified. A run described only as spending \(X\) post-training FLOPs leaves three mechanistically different questions unanswered: how much of \(X\) generated rollouts, how much updated the policy, and how much evaluated rewards or feedback models? These quantities are not bookkeeping details. Different reward systems can appear comparable under total compute while using different mechanisms; PRM-based training, for example, spends part of the budget on an additional model. If that cost is ignored, the apparent policy-update allocation can be misleading.

We study this fixed-budget allocation problem in GRPO post-training of LoRA-adapted language-model policies on mathematical reasoning tasks. These are not embodied-control experiments; rather, they isolate a compute-allocation problem for foundation-model components increasingly used inside robot-learning pipelines. As robot-learning systems increasingly use foundation models for planning, feedback, and reward modeling, RL post-training compute must be allocated not only between rollouts and updates, but also across learned feedback mechanisms. The controlled setting lets us compare reward systems, model sizes, and downstream evaluation targets under explicit FLOP accounting.

This positioning matters for robot learning because the cost of improving a reasoning or feedback module is often paid before deployment, while the resulting module may be reused across tasks, embodiments, or evaluation loops. Compute-constrained robot-learning settings often need to decide whether to use a larger model, a smaller model trained longer, more rollout search, or a stronger feedback model. A post-training recipe that spends most compute on reward-model inference can be preferable under one target and wasteful under another; likewise, a recipe that maximizes native training reward may not maximize final-answer transfer. We therefore treat allocation itself as the object of study, rather than treating RL post-training compute as an undifferentiated scalar.

Our thesis is that fixed-budget RL post-training is a constrained resource-allocation problem over model size, rollout/search, policy updates, and reward/feedback evaluation. The resulting \emph{conditional allocation frontier} depends on model size, compute budget, reward system, and evaluation target. We make exactly four contributions:
\begin{itemize}
    \item We define FLOP accounting for fixed-budget RL post-training across model size, rollout/search, policy-update/learning, and reward-evaluation compute.
    \item We provide same-FLOP model-size comparisons showing that model choice and training allocation are coupled.
    \item We show reward and evaluation-target dependence: native reward, pass@1 symbolic accuracy, and common-judge process scores can select different best observed allocations.
    \item We introduce RACE, a diagnostic pilot-grid protocol for identifying allocation regimes before expensive validation runs.
\end{itemize}

The empirical message is intentionally not that one recipe beats all baselines by a large downstream margin. Instead, we show that the answer to ``where should compute go?'' depends on what is being measured. Model size changes allocation behavior, reward design changes native-reward allocation, and evaluation targets can disagree about the selected frontier.

\section{Related Work}

Scaling laws established that model quality depends predictably on model size, data, and compute, and that compute-efficient training requires balancing scaling axes rather than maximizing one of them \citep{kaplan2020scaling, hoffmann2022training}. Recent RL post-training scaling work extends this perspective to preference optimization and mathematical reasoning, showing that RL performance can improve systematically with additional compute \citep{scalerl2025, rlmathscaling2025}. Our focus is orthogonal: instead of varying total RL compute, we hold compute approximately fixed and study where it should be spent.

Test-time reasoning methods such as chain-of-thought prompting, self-consistency, and tree search show that additional inference compute can improve reasoning quality \citep{wei2022chain, wang2023selfconsistency, yao2023tree}. Related robotics work uses language models as planners, program generators, and embodied reasoning modules \citep{huang2022zeroshot,liang2023code,huang2023inner}. RL post-training couples this search process to policy learning: sampled rollouts explore candidate reasoning traces, and gradient updates convert feedback into parameter changes. This makes rollout count, rollout length, update steps, and trainable parameter count part of one allocation problem.

Reward quality is also a scaling axis. Sparse outcome rewards are cheap but weak; dense rule-based rewards provide more frequent signal; process reward models and learned verifiers can score intermediate reasoning but add inference cost \citep{christiano2017preferences, ouyang2022instructgpt, bai2022helpful, genrm2025}. Robot-learning pipelines face the same trade-off when learned critics, reward models, or evaluators are used to supervise foundation-model components. We explicitly account for this reward-evaluation compute and test whether reward systems shift the best observed allocation frontier.

\section{Method: Conditional Compute Allocation}
\label{sec:compute_framework}

We model a post-training run as the sum of three compute terms:
\begin{equation}
C_{\text{total}} =
C_{\text{search}} +
C_{\text{learning}} +
C_{\text{reward}}.
\end{equation}
Here $C_{\text{search}}$ is autoregressive rollout generation, $C_{\text{learning}}$ is policy optimization, and $C_{\text{reward}}$ is reward or verifier inference. We estimate these terms using parameter--token FLOP accounting and log the realized token counts for every run. For LoRA training, update compute uses an effective update fraction that accounts for both trainable adapter parameters and frozen-backbone forward/activation costs; details are in Appendix~\ref{app:compute_accounting}.

The decomposition is deliberately operational. Rollout FLOPs increase with the number and length of sampled completions; learning FLOPs increase with update steps, trainable rank, and the effective backward cost; reward FLOPs are negligible for rule-based rewards but substantial for PRM-based feedback. This makes two runs with comparable total FLOPs meaningfully different if they allocate those FLOPs to different channels.

The accounting also exposes when comparisons are mechanistically unfair. A rule-based reward run spends almost all non-update compute on policy rollouts, whereas a learned-feedback run may spend a substantial fraction on reward-model inference. Both can have the same \(C_{\text{total}}\), but one budget buys exploration and the other buys feedback. Our goal is to make this trade-off visible.

Our main allocation variable is the total-budget update fraction
\begin{equation}
\rho =
\frac{C_{\mathrm{learning}}}
{C_{\mathrm{search}} + C_{\mathrm{learning}} + C_{\mathrm{reward}}}.
\end{equation}
Small $\rho$ is search- or reward-heavy; large $\rho$ is update-heavy. We use ``best observed $\rho$'' throughout because the experiments estimate conditional frontiers rather than proving globally best allocations.

\paragraph{RACE diagnostic.}
Reward-Aware Compute Allocation (RACE) is a diagnostic protocol for measuring this frontier under limited pilot budgets. For each model, reward system, and compute band, RACE runs a small IsoFLOP pilot grid, estimates the local reward--allocation response, and selects validation allocations near the best observed or fitted regime. If the fit is unstable or boundary-dominated, RACE falls back to pilot-best regime selection. We evaluate RACE as a diagnostic: it should recover useful allocation regimes and identify when pooled rules fail, but it is not claimed to guarantee held-out reward improvement.

This distinction is important. RACE is not a search algorithm over all possible post-training recipes; it is a measurement protocol for asking whether a local compute band is search-heavy, update-heavy, or reward-compute-heavy. Its output is a regime hypothesis that can guide validation, not a proof that a selected run will dominate all alternatives.

\section{Experimental Design}

All experiments use GRPO with LoRA-adapted Qwen2.5 instruction-tuned policies on Polaris-53K mathematical reasoning. We fix $K=2$ sampled completions and maximum completion length $L=2048$ unless stated otherwise, and vary update steps, LoRA rank, model size, and reward system to sweep $\rho$ inside approximate IsoFLOP bands.

\paragraph{Stages.}
Stage A studies model-aware search--learning allocation across 1.5B, 3B, and 7B policies. Stage B fixes the 3B policy and compares sparse, structured, dense, and proxy-PRM rewards. Stage D replaces proxy supervision with a real process reward model, Qwen2.5-Math-PRM-7B. Exploratory verifier-capacity runs are treated as secondary diagnostics and are not part of the main evidence.

The reward systems are chosen to span a practical feedback spectrum. Sparse rewards score only final-answer correctness; structured rewards add formatting and answer-shape constraints; dense rewards add bounded partial credit; proxy PRM rewards approximate process supervision with an inexpensive proxy signal; real PRM rewards use a dedicated 7B process reward model. This ordering increases feedback richness but also changes reward-evaluation cost, which is why reward compute must enter the allocation denominator.

\paragraph{Metrics.}
We report native training reward, min--max normalized reward for cross-reward comparisons, and common final-answer accuracy. For downstream validation, we evaluate saved adapters on GSM8K and MATH-500 using exact final-answer accuracy, symbolic equivalence accuracy, pass@1, and a smaller pass@3 setting when available. We use downstream metrics to test whether reward-conditioned frontiers transfer beyond the training reward.

The downstream evaluation is intentionally separated from native reward selection. Native reward answers whether a reward system prefers a particular allocation under its own feedback signal. Downstream final-answer accuracy asks whether that preference transfers to held-out reasoning benchmarks. The common judge asks a third question: whether a shared process-quality evaluator prefers the same frontier. Disagreement across these targets is informative rather than a failure mode.

\section{Results}

\paragraph{RQ1: Does fixed-budget RL post-training have structured allocation behavior?}
Stage A shows a structured allocation plane under matched post-training budgets. Search-heavy settings underuse the available learning capacity, while update-heavy settings can become exploration-limited. The resulting frontier is visible only when rollout, update, and reward-evaluation FLOPs are separated rather than collapsed into one aggregate budget.

\begin{figure}[t]
    \centering
    \includegraphics[width=\linewidth]{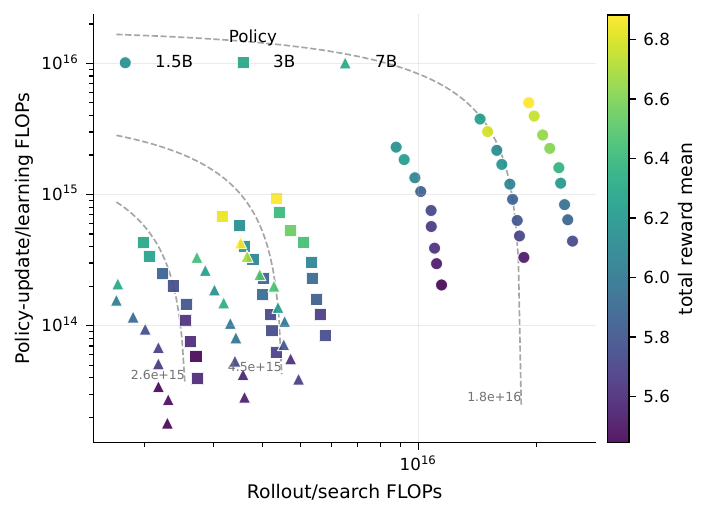}
    \caption{\textbf{Search--learning allocation plane.} Under a fixed RL post-training budget, runs differ in how much compute is spent on rollout search, policy updates, and reward evaluation. The best observed allocation is a point on this conditional resource-allocation plane, not a function of total compute alone.}
    \label{fig:allocation_plane}
\end{figure}

Figure~\ref{fig:allocation_plane} visualizes this directly: runs with comparable total FLOPs occupy different locations in the search--learning plane, and their reward varies with that location. The important observation is not a single best point, but the existence of a measurable response surface. This is the allocation analogue of a scaling curve: the independent variable is not more compute, but how a fixed budget is divided.

\begin{figure}
    \centering
    \includegraphics[width=\linewidth]{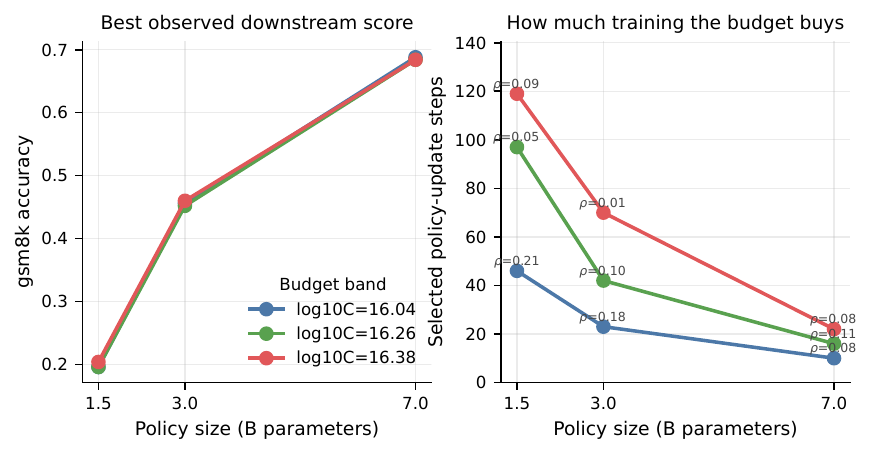}
    \caption{\textbf{Same-FLOP model-size trade-off.} Under matched pilot budgets, changing policy size changes how much training the budget buys. Larger policies consume more per-token compute, so the same FLOP budget can imply fewer rollout tokens and fewer update steps; the best observed downstream run therefore depends jointly on model size and allocation.}
    \label{fig:model_conditioned_frontier}
\end{figure}

\paragraph{RQ2: Are allocation frontiers model- and reward-conditioned?}
Pooling across model sizes obscures the allocation response: under the same FLOP budget, a larger policy consumes more compute per token, so the budget buys fewer rollouts and fewer update steps. Figure~\ref{fig:model_conditioned_frontier} makes this fixed-resource trade-off explicit: model choice and training allocation are coupled, rather than separable decisions. Reward design also shifts the frontier. In Stage B/D, sparse, structured, and dense rule-based rewards often select high-update regimes around $\rho\approx0.72$, while proxy and real PRM objectives shift the native-reward frontier toward lower update fractions around $\rho\approx0.44$--$0.45$ in the matched 3B setting once reward-model compute is counted. We do not interpret this as a fixed rule for PRMs; the PRM frontier is itself model-size and target dependent.

This result is the main empirical reason to avoid pooled allocation prescriptions. A smaller policy trained longer, a larger policy trained for fewer steps, a cheap rule-based reward, and an expensive learned feedback model can all consume the same total budget while needing different trade-offs between search and learning. In this sense, model size and reward design are both part of the scaling problem: they change both the signal being optimized and the compute path used to obtain that signal.

The PRM result should be read through this accounting lens. The lower native-reward update fraction is not evidence that PRMs intrinsically require less learning. It reflects that learned feedback consumes part of the fixed budget, so the same total-compute number corresponds to a different mixture of rollout, update, and reward-evaluation work.

\begin{figure}
    \centering
    \includegraphics[width=\linewidth]{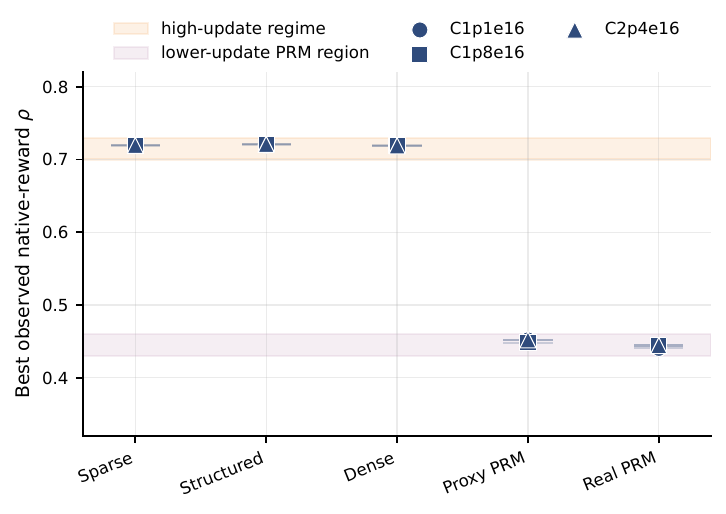}
    \caption{\textbf{Feedback cost changes allocation.} For best observed native-reward runs in the matched 3B setting, rule-based rewards spend the budget on policy rollout and update, while PRM-style feedback allocates a visible fraction to reward-model evaluation. This is the mechanism hidden by reporting only total FLOPs.}
    \label{fig:reward_vs_downstream_frontier}
\end{figure}

\paragraph{RQ3: Do downstream evaluation targets select different allocations?}
We evaluate saved adapters on GSM8K and MATH-500 using 250 examples per dataset. Downstream pass@1 symbolic accuracy often favors the high-update a180 setting; the real-PRM a180 run has the best listed average pass@1 accuracy, approximately 0.714 across GSM8K and MATH-500. By contrast, the partial common-judge evaluation often favors the mid-update a100 setting. The preferred allocation therefore depends on the evaluation target: native reward, downstream final-answer accuracy, and process-quality scores can select different frontiers.

  \begin{table}[t]
  \centering
  \small
  \caption{\textbf{Downstream target dependence under the matched 3B setting.} Pass@1 is symbolic-equivalence accuracy averaged over
  GSM8K and MATH-500 with 250 examples per dataset. Common judge is a shared process-quality score evaluated for dense and PRM-style
  rewards. Native $\rho$ reports the best observed native-reward allocation in the matched Stage B/D grid.}
  \label{tab:target_dependence}
  \begin{tabular}{lccccc}
  \toprule
  Reward & Native $\rho$ & Pass@1 best & Judge a40 & Judge a100 & Judge a180 \\
  \midrule
  Sparse & 0.719 & a180: \textbf{0.708} & -- & -- & -- \\
  Structured & 0.720 & a40/a180: \textbf{0.698} & -- & -- & -- \\
  Dense & 0.719 & a180: \textbf{0.702} & 0.757 & \textbf{0.780} & 0.759 \\
  Proxy PRM & 0.448--0.452 & a180: \textbf{0.708} & 0.746 & \textbf{0.767} & 0.749 \\
  Real PRM & 0.440--0.444 & a180: \textbf{0.714} & 0.743 & \textbf{0.761} & 0.746 \\
  \bottomrule
  \end{tabular}
  \end{table}
Table~\ref{tab:target_dependence} makes the target dependence explicit. Native reward separates rule-based and PRM-style feedback into different allocation regimes. Pass@1 accuracy, however, often favors the high-update a180 setting across reward systems. The common judge, where available, instead peaks at a100 for dense, proxy PRM, and real PRM. This does not make one metric more valid than the others; it shows that allocation decisions must name the target they are intended to optimize.

\begin{figure}[t]
    \centering
    \includegraphics[width=\linewidth]{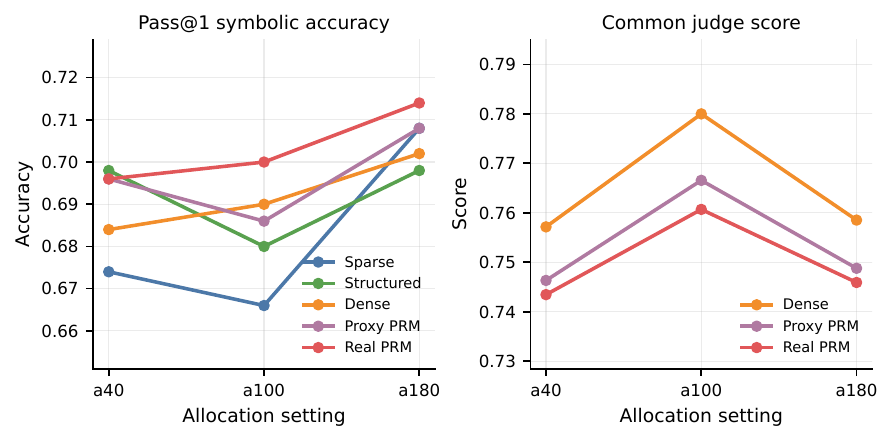}
    \caption{\textbf{Evaluation target changes allocation.} Native reward, downstream pass@1 symbolic accuracy, and common process-quality scoring can select different allocation regimes. The answer to ``where should compute go?'' therefore depends on the target being measured.}
    \label{fig:target_dependence_pass1_vs_judge}
\end{figure}

\paragraph{RQ4: Can RACE diagnose allocation regimes from small pilot grids?}
RACE recovers the correct high- versus lower-update regime in inside-grid and leave-rank-out validation, with same-regime rates of 1.0 in both settings. However, reward regret is nonzero, and true held-out validation does not improve the global native-reward baseline. We therefore present RACE as a diagnostic protocol for selecting allocation regimes and prioritizing validation runs, not as a method that guarantees held-out improvement.

The mixed held-out behavior is useful for the paper's claim. It suggests that small pilot grids can identify whether a setting belongs to a high-update or lower-update regime, but local curve fits are not sufficient to certify a globally better recipe. For practical RL post-training, this means RACE should be used to reduce the number of expensive validation runs and to expose conditional structure, while final claims should still be based on held-out downstream evaluation.

\section{Conclusion}
\label{sec:conclusion}

RL post-training under a fixed budget is not characterized by total compute alone. Across model sizes and reward systems, we observe conditional allocation frontiers: the best observed update/search/reward-compute balance depends on model size, reward design, and evaluation target. This suggests that practical post-training recipes for foundation models, including components used in robot-learning pipelines, should report and tune compute allocation rather than only aggregate FLOPs.

In practice, the framework helps decide whether limited post-training compute should be spent on a larger policy, more updates to a smaller policy, more rollout search, or more expensive feedback. This is the resource-allocation problem faced by RL post-training pipelines: the same FLOP budget can buy different mixtures of model capacity, sampled experience, gradient updates, and reward-model inference.

For robot-learning systems that use foundation models as planners, feedback interpreters, reward models, or evaluators, this has a concrete implication: improving a module by RL post-training is itself a resource-allocation problem. Reporting the total post-training budget without its rollout, update, and feedback decomposition can hide why a recipe transfers, fails, or changes behavior under a different evaluation target. We therefore recommend reporting allocation-aware compute: total FLOPs together with rollout FLOPs, update FLOPs, reward-evaluation FLOPs, and the resulting update fraction \(\rho\).

\section{Limitations}

Our experiments use mathematical reasoning as a controlled proxy for foundation-model post-training rather than direct embodied robot tasks. The results should therefore be interpreted as evidence about RL post-training of reasoning and feedback components used in robot-learning pipelines, not as a claim about embodied policy learning. We use one model family, GRPO, LoRA adaptation, and limited seeds. FLOP estimates are designed for consistent relative comparisons and do not model all hardware effects. RACE is a regime-selection diagnostic: it identifies useful allocation regions but does not guarantee reward improvement in held-out runs. Exploratory judge-compute experiments are not part of the main evidence. We do not claim a fixed best $\rho$ or a general post-training law across tasks, models, and feedback systems.



\clearpage


\bibliography{example}  

\appendix

\section{Experimental Details}
\label{app:experimental_details}

\subsection{Overview of Experimental Stages}

We organize the experiments into three main empirical stages. Stage A studies
the core search--learning allocation frontier under an IsoFLOP design. Stage B
extends the same compute-allocation protocol to compare reward systems. Stage D
replaces proxy reward supervision with a dedicated process reward model (PRM).

Stage A uses the original IsoFLOP model-scaling design. It spans multiple model
sizes, compute bands, and LoRA ranks. Stage B fixes the policy backbone and
varies the reward system under matched compute. Stage D uses the same fixed
policy backbone but replaces the proxy reward with a dedicated PRM-based reward
model.

\subsection{Stage A: Model-Aware IsoFLOP Design}

Stage A uses instruction-tuned Qwen2.5 models with sizes
\[
N \in \{1.5\mathrm{B}, 3\mathrm{B}, 7\mathrm{B}\}.
\]
For all Stage A runs, we use \(K=2\) sampled completions per prompt and a
maximum completion length of \(L=2048\). The initial LoRA ranks are
\[
r \in \{8,16,32,64,128\},
\]
with additional intermediate ranks
\[
r \in \{12,24,48,96\}
\]
added to densify the IsoFLOP frontier.

The three compute bands are denoted
\[
\texttt{C1p1e16},\quad \texttt{C1p8e16},\quad \texttt{C2p4e16}.
\]
For the 1.5B reference model, the base step schedules are:
\[
\begin{aligned}
\texttt{C1p1e16}:&\quad r8=65,\ r16=62,\ r32=59,\ r64=53,\ r128=46,\\
\texttt{C1p8e16}:&\quad r8=105,\ r16=100,\ r32=95,\ r64=85,\ r128=74,\\
\texttt{C2p4e16}:&\quad r8=140,\ r16=133,\ r32=127,\ r64=113,\ r128=99.
\end{aligned}
\]
Intermediate ranks are obtained by log-rank interpolation. For other model
sizes, we scale the number of update steps as
\[
S(N) = \mathrm{round}(S_{1.5\mathrm{B}} \cdot 1.5/N_B),
\]
where \(N_B\) is the model size in billions of parameters.

\subsection{Stage B: Reward-Conditioned DOE}

Stage B studies how reward systems change the allocation frontier. We fix the
policy backbone to
\[
\texttt{Qwen/Qwen2.5-3B-Instruct}
\]
and use \(K=2\), \(L=2048\), and the same three compute bands as Stage A. The
full Stage B design uses
\[
r \in \{8,12,16,24,32,48,64,96,128\}
\]
and compares four reward systems:
\[
\mathcal{R} \in \{\text{sparse},\text{structured},\text{dense},\text{proxy PRM}\}.
\]
Thus the full Stage B design contains
\[
3 \text{ compute bands} \times 9 \text{ ranks} \times 4 \text{ reward systems}
= 108
\]
runs.

The sparse reward uses binary final-answer correctness. The structured reward
combines final-answer correctness with formatting and answer-shape terms. The
dense reward provides bounded partial credit based on numerical closeness and
format compliance. The proxy PRM reward combines outcome supervision with a
process-style reward signal.

\subsection{Stage D: Real PRM Validation}

Stage D evaluates dedicated process reward modeling. We again fix the policy
backbone to
\[
\texttt{Qwen/Qwen2.5-3B-Instruct},
\]
use \(K=2\), \(L=2048\), and the same three compute bands and LoRA ranks as
Stage B. The reward system is a real process reward model based on
\[
\texttt{Qwen/Qwen2.5-Math-PRM-7B}.
\]
The full Stage D design contains
\[
3 \text{ compute bands} \times 9 \text{ ranks} = 27
\]
runs.

We refer to this experiment as real PRM validation rather than verifier-capacity scaling,
since the PRM model size is fixed.

\section{Compute Accounting}
\label{app:compute_accounting}

We decompose total post-training compute into rollout/search compute, policy
update compute, and reward-model compute:
\[
F_{\mathrm{total}}
=
F_{\mathrm{rollout}}
+
F_{\mathrm{update}}
+
F_{\mathrm{reward}}.
\]

Rollout compute is estimated as
\[
F_{\mathrm{rollout}}
=
c_{\mathrm{fwd}}(N+R)T_{\mathrm{rollout}},
\]
where \(N\) is the policy backbone parameter count, \(R\) is the reward-model
parameter count when a learned reward model is used, and \(T_{\mathrm{rollout}}\)
is the number of rollout tokens.

Update compute is estimated as
\[
F_{\mathrm{update}}
=
c_{\mathrm{bwd}}(N f_{\mathrm{eff}})T_{\mathrm{update}},
\]
where \(T_{\mathrm{update}}\) is the number of tokens used for learning. For
LoRA-based training, we define
\[
f_{\mathrm{LoRA}} = \frac{N_{\mathrm{LoRA}}}{N}
\]
and use an effective update fraction
\[
f_{\mathrm{eff}}
=
a + (1-a)f_{\mathrm{LoRA}}.
\]
The constant \(a\) accounts for update-side costs that depend on the frozen
backbone, including forward passes, activations, and shared computation. Unless
otherwise stated, we use \(a=0.85\).

We summarize compute allocation using the update fraction
\[
\rho =
\frac{F_{\mathrm{update}}}{F_{\mathrm{total}}}.
\]
Small \(\rho\) corresponds to search- or reward-heavy regimes, while large
\(\rho\) corresponds to update-heavy regimes.

\section{Metrics and Reward Comparability}
\label{app:metrics}

Different reward systems use different native reward scales. Therefore, raw
reward values are not directly comparable across sparse, structured, dense,
proxy PRM, and real PRM conditions. We report three complementary metrics.

First, we report the native reward
\[
R_{\mathrm{native}},
\]
which is the scalar reward optimized by the training run.

Second, for cross-reward comparisons, we compute normalized reward within each
reward system and compute band:
\[
R_{\mathrm{norm}}
=
\frac{R - \min_{\mathcal{R},C} R}
{\max_{\mathcal{R},C} R - \min_{\mathcal{R},C} R}.
\]
This preserves within-reward ordering while avoiding direct comparison of
incompatible native reward scales.

Third, we report a common accuracy metric when available. In our analysis code,
this is stored as \texttt{accuracy\_main}. When exact held-out validation
accuracy is unavailable, this quantity should be interpreted as a common
correctness proxy rather than a calibrated benchmark score.

\section{Reward Definitions}
\label{app:reward_definitions}

The sparse reward is binary:
\[
r_{\mathrm{sparse}} = \mathbb{1}[\hat{y}=y^\star],
\]
where \(\hat{y}\) is the extracted model answer and \(y^\star\) is the target
answer after symbolic normalization.

The structured reward combines final-answer correctness with formatting and
answer-shape constraints:
\[
r_{\mathrm{structured}}
=
r_{\mathrm{answer}}
+
\lambda_{\mathrm{fmt}} r_{\mathrm{format}}
+
\lambda_{\mathrm{shape}} r_{\mathrm{shape}}.
\]

The dense reward provides bounded partial credit:
\[
r_{\mathrm{dense}} \in [-1,1],
\]
combining symbolic correctness, numerical proximity, and output-format
constraints.

The proxy PRM reward combines process-style supervision and outcome
correctness:
\[
r_{\mathrm{proxy\ PRM}}
=
\alpha r_{\mathrm{proc}}
+
(1-\alpha)r_{\mathrm{out}},
\]
where \(r_{\mathrm{proc}}\) is an intermediate process score and
\(r_{\mathrm{out}}\) is final-answer correctness.

The real PRM reward uses a dedicated process reward model to score intermediate
reasoning steps. In Stage D, this model is
\[
\texttt{Qwen/Qwen2.5-Math-PRM-7B}.
\]

\section{RACE: Reward-Aware Compute Allocation}
\label{app:race}

RACE is a diagnostic protocol for selecting targeted post-training validation
runs from an initial IsoFLOP pilot grid. Given a fixed model size \(N\), reward
system \(\mathcal{R}\), and compute band \(C\), RACE fits a local response
model
\[
\hat{R}_{N,\mathcal{R},C}(\rho)
\]
over the update fraction \(\rho\). The default response model is a quadratic in
log update fraction:
\[
\hat{R}(\rho)
=
a(\log \rho)^2 + b\log \rho + c.
\]

RACE uses this fit only as a decision rule. If the fitted best-response point is stable and
lies inside the observed range, RACE recommends validation runs near the
predicted point. If the best-response point lies near a boundary, RACE recommends
held-out runs beyond the corresponding boundary. If the fit is unstable, RACE
falls back to conservative validation near the best observed pilot run.

Importantly, RACE is applied conditionally. We do not fit one pooled curve
across all model sizes and reward systems. Instead, the response is conditioned
on model size, compute band, and reward system. We write this conditional
regime estimate schematically as
\[
\hat{\rho}_{\mathrm{RACE}} = \hat{\rho}_{\mathrm{RACE}}(N,C,\mathcal{R}).
\]

\section{RACE Validation Protocol}
\label{app:race_validation}

The Stage A RACE validation uses the compute band \texttt{C1p8e16}. The initial
grid contains ranks
\[
r \in \{8,12,16,24,32,48,64,96,128\}.
\]
Held-out validation uses ranks
\[
r \in \{160,192,256\}.
\]
The validation grid includes model sizes \(1.5\mathrm{B}\), \(3\mathrm{B}\),
and \(7\mathrm{B}\), all with \(K=2\) and \(L=2048\).

The validation compares the best held-out configuration against the best
in-grid configuration in the same compute band. In the current Stage A
validation, the best held-out configuration improves some model-conditional
subgroups and improves accuracy for the best 7B validation run, but it does not
improve the primary native reward over the best global in-grid baseline.
Therefore, we report RACE as a conservative diagnostic protocol rather than as a
method that guarantees held-out improvement.

\section{RACE Fit-Model Ablation}
\label{app:race_fit_ablation}

RACE uses a quadratic response model in $\log \rho$ as a transparent decision
model for selecting held-out validation configurations. To test whether this
choice is statistically preferred on the Stage A pilot grid, we ablate several
low-capacity response families: a constant mean baseline, linear and quadratic
models in $\rho$, linear and quadratic models in $\log \rho$, and a cubic model
in $\log \rho$.

Models are fit independently within each compute band and compared using
leave-one-out cross-validation RMSE, mean absolute error, and BIC. On the
current Stage A data, the constant baseline obtains the lowest average LOOCV
RMSE, while the linear log-rho model is the best nontrivial allocation-dependent
model. This indicates that the pooled Stage A response is noisy and partially
confounded by model size and rank variation.

We therefore do not interpret the quadratic log-rho fit as the statistically
preferred response model. Instead, we use it as a transparent boundary diagnostic:
it identifies whether the pilot grid suggests an interior best-response region, a boundary
best-response region, or an unstable fit. When the fit is unstable or boundary-dominated,
RACE falls back to conservative held-out validation rather than trusting an
extrapolated point. This conservative behavior is central to the method.

\section{RACE Implementation Details}
\label{app:race_details}

The initial Stage A grid uses LoRA ranks
\[
r \in \{8,12,16,24,32,48,64,96,128\}.
\]
Held-out RACE validation runs use ranks
\[
r \in \{160,192,256\},
\]
which are excluded from the pilot fit. The initial compute bands are
\texttt{C1p1e16}, \texttt{C1p8e16}, and \texttt{C2p4e16}. For the 1.5B reference
model, the base step schedules are:
\[
\begin{aligned}
\texttt{C1p1e16}:&\quad r8=65,\ r16=62,\ r32=59,\ r64=53,\ r128=46,\\
\texttt{C1p8e16}:&\quad r8=105,\ r16=100,\ r32=95,\ r64=85,\ r128=74,\\
\texttt{C2p4e16}:&\quad r8=140,\ r16=133,\ r32=127,\ r64=113,\ r128=99.
\end{aligned}
\]
For other model sizes, steps are scaled as
\[
S(N) = \mathrm{round}(S_{1.5B} \cdot 1.5 / N_B),
\]
where $N_B$ is the model size in billions of parameters.

Validation ranks are selected by log-rank interpolation or extrapolation of the
same schedules. Runs are labeled as validation if their rank is at least 160 or
if their path contains one of the validation keywords:
\texttt{reuse}, \texttt{prefill}, \texttt{validation}, \texttt{val},
\texttt{extra}, \texttt{target}, \texttt{heldout}, or \texttt{race}.

\end{document}